\documentclass{article}

\usepackage[preprint]{neurips_2026}

\usepackage[utf8]{inputenc}
\usepackage[T1]{fontenc}

\usepackage{float}
\usepackage{graphicx}
\usepackage{booktabs}
\usepackage{multirow}
\usepackage{amsmath}
\usepackage{amsfonts}
\usepackage{amssymb}
\usepackage{microtype}
\usepackage{url}
\usepackage{hyperref}

\newcommand{\Kstar}{K_{\star}}
\newcommand{\Kgene}{K_{\mathrm{gene}}}
\newcommand{\Kspat}{K_{\mathrm{spat}}}
\newcommand{\Keval}{K_{\mathrm{eval}}}

\newcommand{\Lret}{\mathcal{L}_{\mathrm{ret}}}
\newcommand{\Lsoft}{\mathcal{L}_{\mathrm{soft}}}
\newcommand{\Lglob}{\mathcal{L}_{\mathrm{glob}}}
\newcommand{\Lloc}{\mathcal{L}_{\mathrm{loc}}}

\title{
BioKERN: Biological Kernel Regularization for
Histology-to-Transcriptomics Neighborhood Retrieval
}

\author{%
  Seungik Cho \\
  Department of Physics and Astronomy\\
  Rice University\\
  Houston, TX 77005\\
  \texttt{sc252@rice.edu}
  \And
  Betul Orcan-Ekmekci\thanks{Corresponding author.} \\
  Department of Mathematics\\
  Rice University\\
  Houston, TX 77005 \\
  \texttt{orcan@rice.edu}
}

\begin{document}

\maketitle


\begin{abstract}
Spatially resolved biology requires representations that preserve biological
neighborhood structure rather than only exact cross-modal correspondences.
Existing histology--transcriptomics objectives can emphasize instance-level
matching even when non-paired spots share molecular or spatial context. We
introduce \textbf{BioKERN}, a multimodal spatial representation-learning
framework that incorporates biological structure as an explicit, learnable
inductive bias. BioKERN constructs a training-time biological kernel by
combining transcriptomic similarity and spatial proximity, then uses it to
provide graded neighborhood supervision and regularize embedding geometry.
Evaluation uses a fixed, model-independent biological neighborhood definition
shared by all methods. Across Mouse Brain Visium and Human Liver GSE240429,
BioKERN consistently improves biological-neighborhood retrieval over BLEEP in
both single- and multi-scale settings. Controlled shared-architecture
experiments show that most of the improvement arises from biological-kernel
regularization rather than increased model capacity. These results support
explicit biological geometry as an interpretable inductive bias for multimodal
learning in spatial biology.
\end{abstract}


\section{Introduction}
\label{sec:introduction}

Spatial transcriptomics measures molecular state while preserving tissue
location, providing a direct view of how gene-expression programs are
organized within anatomical structure
~\citep{stahl2016visualization,stickels2021highly,chen2022spatiotemporal}.
Histology provides a complementary morphological view and is available at
substantially larger scale in both research and clinical workflows
~\citep{campanella2019clinical,lu2021data}. This asymmetry motivates a natural
multimodal learning problem: given an H\&E patch from a new tissue, retrieve
transcriptomic profiles representing the same biological context from a
profiled reference.

Exact image--spot correspondence is useful supervision, but it does not fully
capture the organization of spatial tissue. Spatially adjacent observations,
spots within the same tissue domain, or locations with similar transcriptional
programs can represent related biological states even when they are not the
same measured location. A representation optimized only for instance identity
can therefore under-represent biologically meaningful neighborhood structure.

Biological similarity is also multi-factorial. Transcriptomic similarity
reflects molecular programs, whereas spatial proximity reflects tissue
organization~\citep{palla2022squidpy,dries2021giotto}. Their relative utility
can differ between tissue settings, motivating a learnable rather than fixed
combination of these signals.

We introduce \textbf{BioKERN}
(\textbf{Bio}logical \textbf{KE}rnel
\textbf{R}egularization for histology-to-transcriptomics
\textbf{N}eighborhood retrieval), which incorporates explicit biological
geometry into multimodal spatial representation learning. BioKERN constructs
a reference kernel from transcriptomic similarity and spatial proximity,
learns their relative weighting, and regularizes the H\&E--transcriptomics
representation to preserve this geometry.

We formulate the resulting task as \emph{biological neighborhood retrieval}
(BNR): given an H\&E query, the representation should retrieve
transcriptomic states belonging to the same biological neighborhood rather
than being evaluated only by exact-pair matching.

Our main contributions are as follows.
\begin{itemize}
    \item \textbf{Biologically structured multimodal representation learning.}
    BioKERN augments instance-level alignment with graded supervision derived
    from explicit biological neighborhood similarity.

    \item \textbf{Learnable molecular--spatial weighting.}
    BioKERN combines transcriptomic similarity and spatial proximity through a
    learned scalar weight, yielding a compact and interpretable biological
    prior.

    \item \textbf{Controlled validation of the biological prior.}
    Shared-architecture controls, shuffled-kernel controls, and ablations show
    that biological-kernel regularization accounts for most of the improvement
    in biological-neighborhood retrieval.
\end{itemize}


\section{Related Work}
\label{sec:related}

\paragraph{Histology--transcriptomics representation learning.}
BLEEP~\citep{xie2023bleep} learns a joint image--gene representation for
retrieval-based expression prediction and uses similarity-smoothed targets to
reduce purely instance-discriminative supervision. Other multimodal or spatial
methods, including mclSTExp~\citep{yang2024mclstexp},
ConGcR~\citep{zhou2024congcr}, SpaGCN~\citep{hu2021spagcn},
STAGATE~\citep{dong2022stagate}, and GraphST~\citep{long2023graphst},
incorporate spatial context or graph structure. BioKERN differs by defining an
explicit biological target geometry from transcriptomic and spatial
similarity, then using that geometry for both graded retrieval supervision and
direct global/local representation regularization.

\paragraph{Structured relational supervision.}
Kernel alignment~\citep{cristianini2002kernel,cortes2012algorithms},
relational representation learning
~\citep{park2019relational,tung2019similarity}, and supervised contrastive
learning~\citep{khosla2020supervised} show that pairwise structure can guide
representations beyond one-hot instance labels. BioKERN adapts this principle
to spatial biology using a learnable molecular--spatial reference kernel.
Extended related work is provided in Appendix~\ref{app:related}.


\section{Method}
\label{sec:method}

\begin{figure}[t]
    \centering
    \includegraphics[width=\linewidth]{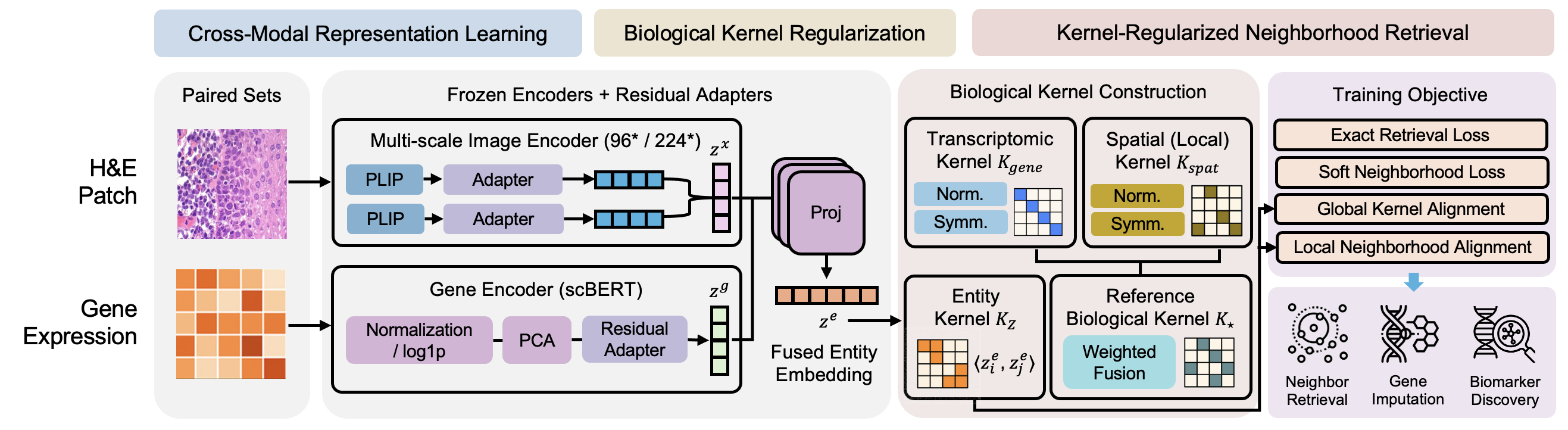}
    \caption{
    \textbf{BioKERN overview.}
    Multimodal image and gene representations are regularized during training
    using a biological reference kernel combining transcriptomic and spatial
    similarity. The kernel defines graded neighborhoods and supervises global
    and local embedding geometry; inference requires only the learned image and
    transcriptomic embeddings.
    }
    \label{fig:overview}
\end{figure}

BioKERN learns a multimodal spatial representation in which cross-modal
similarity reflects paired image--gene correspondence together with biological
neighborhood structure. The framework consists of multimodal representation
learning, construction of a learnable biological reference kernel, and
kernel-regularized neighborhood supervision
(Figure~\ref{fig:overview}).

\paragraph{Multimodal spatial representations.}
For each spatial-transcriptomics spot, we observe
$(x_i,g_i,s_i)$, where $x_i$ is an H\&E image patch,
$g_i$ is the corresponding gene-expression profile, and
$s_i\in\mathbb{R}^2$ is its tissue coordinate.
A frozen PLIP pathology encoder~\citep{huang2023visual} extracts morphology
features, while gene expression is normalized, log-transformed, standardized,
and reduced with PCA. Lightweight ResidualAdapters project both modalities
into a shared $d=128$ dimensional $\ell_2$-normalized embedding space.
We evaluate two image-context settings: the \textbf{single-scale (SS)}
setting uses a $96\times96$ H\&E patch to capture local morphology, whereas
the \textbf{multi-scale (MS)} setting combines representations from
$96\times96$ and $224\times224$ patches using a learnable fusion weight,
thereby incorporating both local morphology and broader tissue context.
All backbone encoders remain frozen.

\paragraph{Learnable biological reference kernel.}
We describe biological relatedness using complementary transcriptomic and
spatial RBF kernels,
\begin{equation}
\Kgene(i,j)
=
\exp\!\left(
-\frac{\|h_i^g-h_j^g\|_2^2}{2\sigma_g^2}
\right),
\qquad
\Kspat(i,j)
=
\exp\!\left(
-\frac{\|s_i-s_j\|_2^2}{2\sigma_s^2}
\right).
\label{eq:kernels}
\end{equation}
Bandwidths are determined from training-set pairwise distances. Rather than
fixing the relative weighting of molecular and spatial information, BioKERN
learns their composition:
\begin{equation}
\Kstar
=
\alpha\Kgene+(1-\alpha)\Kspat,
\qquad
\alpha=\sigma(a)
\label{eq:biological_kernel}
\end{equation}
where $\alpha$ is optimized end-to-end. For multi-section training, spatial
affinities are defined only within the same tissue section; pairs from
different sections have $\Kspat(i,j)=0$, while transcriptomic affinities remain
available across sections. The resulting scalar gives a compact, interpretable
weighting of the two sources of biological neighborhood structure.

\paragraph{Biological-neighborhood supervision.}
Standard bidirectional InfoNCE preserves exact image--gene correspondence,
\begin{equation}
\Lret
=
\frac{1}{2}
\left[
\mathrm{CE}(S,I)
+
\mathrm{CE}(S^\top,I)
\right],
\qquad
S=Z^x(Z^g)^\top/\tau.
\label{eq:retrieval}
\end{equation}
BioKERN augments this objective with three biologically structured terms.
For each observation, the \textbf{soft neighborhood loss} selects the
minibatch top-$k$ non-self neighbors under $\Kstar$ ($k=20$) and normalizes
their kernel affinities into graded targets. The exact paired observation is
excluded from this biological-neighborhood set, so $\Lsoft$ adds supervision
for distinct but biologically related spots while $\Lret$ retains exact-pair
alignment.

To regularize representation geometry, image and gene embeddings are fused
into a spot-level entity representation $z_i^e$, yielding
$K_Z(i,j)=\langle z_i^e,z_j^e\rangle$. The \textbf{global alignment loss}
matches the overall entity-kernel geometry to $\Kstar$, whereas the
\textbf{local alignment loss} focuses on the strongest biological
neighborhood edges. The complete objective is
\begin{equation}
\mathcal{L}
=
\Lret
+
\lambda_s\Lsoft
+
\lambda_g\Lglob
+
\lambda_\ell\Lloc,
\label{eq:objective}
\end{equation}
where $\lambda_s=0.3$, $\lambda_g=0.1$, and $\lambda_\ell=0.5$.
The biological kernel is a training-time supervisory signal; at inference,
an H\&E query is ranked directly against reference transcriptomic embeddings
by cosine similarity. Detailed loss definitions are provided in
Appendix~\ref{app:method}.


\section{Results}
\label{sec:results}

\paragraph{Experimental setup.}
We evaluate on two 10$\times$ Visium benchmarks with distinct tissue
organization. \textbf{Mouse Brain Visium} uses 2,200 spots with a
1,650/550 train/test split. \textbf{Human Liver GSE240429} uses a
cross-slice transfer setting: A1+B1+D1 (6,963 spots) are used for training and
C1 (2,273 spots) for testing. Single-scale (SS) experiments use a
$96\times96$ H\&E patch, while multi-scale (MS) experiments combine
$96\times96$ and $224\times224$ representations with learned fusion.

For every held-out H\&E query, the retrieval gallery consists of the
transcriptomic profiles from the same held-out test set. The exact paired
profile remains in the gallery for exact-pair metrics but is excluded when
defining biological-neighborhood positives. Bio-mAP uses one fixed,
model-independent evaluation kernel,
$\Keval=0.5\Kgene+0.5\Kspat$, for every method; query-side expression is used
only to construct evaluation labels and is never provided to the retrieval
model. We additionally report transcriptomic-neighbor recall, spatial-neighbor
recall, cluster consistency, and exact-pair retrieval in
Appendix~\ref{app:evaluation}. We compare against CCA, Ridge regression, a
naive PLIP zero-shot lower bound, PLIP linear, and BLEEP. \textbf{BLEEP$^*$}
uses the same ResidualAdapter backbone as BioKERN while retaining BLEEP's
training objective, isolating architectural effects.

\begin{table}[t]
\centering
\scriptsize
\caption{
\textbf{Biological-neighborhood retrieval across tissues and scales.}
Bio-mAP uses the same fixed $\Keval$ for every method; stochastic methods are
mean$\pm$standard deviation over five seeds.
}
\label{tab:main_summary}
\resizebox{\linewidth}{!}{
\begin{tabular}{lccccc}
\toprule
Dataset / Scale
& Ridge
& PLIP linear
& BLEEP
& BLEEP$^*$
& \textbf{BioKERN}
\\
\midrule
Mouse Brain SS
& 0.4902
& 0.5407$\pm$.0033
& 0.5087$\pm$.0033
& 0.5327$\pm$.0036
& \textbf{0.6190$\pm$.0023}
\\
Mouse Brain MS
& 0.5567
& 0.5831$\pm$.0023
& 0.4960$\pm$.0056
& 0.5541$\pm$.0040
& \textbf{0.6716$\pm$.0021}
\\
Human Liver SS
& 0.0383
& 0.0307$\pm$.0001
& 0.0302$\pm$.0005
& 0.0309$\pm$.0004
& \textbf{0.0396$\pm$.0007}
\\
Human Liver MS
& 0.0403
& 0.0319$\pm$.0005
& 0.0312$\pm$.0005
& 0.0321$\pm$.0005
& \textbf{0.0408$\pm$.0005}
\\
\bottomrule
\end{tabular}
}
\end{table}

\paragraph{BioKERN improves biological-neighborhood retrieval.}
BioKERN achieves the highest Bio-mAP among the baselines shown in
Table~\ref{tab:main_summary} across both tissues and image-context settings.
Relative to BLEEP, Mouse Brain Bio-mAP increases from 0.5087 to 0.6190 in SS
and from 0.4960 to 0.6716 in MS. BioKERN also exceeds BLEEP$^*$, indicating
that the gain is not explained by the ResidualAdapter architecture alone. The
Human Liver gains are smaller in absolute magnitude, but persist in the
cross-slice transfer setting and remain competitive with the strong Ridge
baseline.

\paragraph{Biological regularization accounts for most of the gain.}
We decompose the improvement sequentially as
BLEEP $\rightarrow$ BLEEP$^*$ (architecture), BLEEP$^*$ $\rightarrow$
Ret-only (objective change), and Ret-only $\rightarrow$ BioKERN
(biological regularization). Under this additive decomposition, the
biological-regularization step accounts for approximately 63--91\% of the
total Bio-mAP improvement across the four settings. On Mouse Brain SS, the
architecture contributes $+0.0240$, the objective change $+0.0037$, and the
biological-regularization step $+0.0826$. A shuffled-kernel control further
shows that meaningful biological pairwise structure is important rather than
merely adding another relational loss.

\paragraph{Graded neighborhood supervision is the strongest component.}
Ablations show that removing biological-kernel supervision substantially
reduces Bio-mAP, while removing $\Lsoft$ causes the largest degradation among
the three biological objectives. The relative value of molecular and spatial
structure varies across tissue settings: combining both signals performs best
on Mouse Brain and Human Liver MS, whereas Human Liver SS favors the
transcriptomic-only kernel. Consistent with this tissue-dependent behavior,
the learned molecular--spatial weight is stable across multiple
initializations, with Human Liver assigning greater weight to transcriptomic
similarity than Mouse Brain. The strong fixed-$\alpha$ control nevertheless
shows that the primary gain arises from biological-neighborhood supervision
rather than from learning $\alpha$ alone. Full ablations and kernel-weight
analyses are provided in Appendices~\ref{app:ablations}
and~\ref{app:sensitivity}.


\section{Conclusion}
\label{sec:conclusion}

BioKERN introduces an explicit biological inductive bias for multimodal
spatial representation learning. Rather than evaluating representation quality
only through exact-pair matching, BioKERN uses a learnable molecular--spatial
reference geometry to provide graded neighborhood supervision and relational
regularization. Across mouse brain and human liver, this improves
biological-context retrieval, and controlled experiments indicate that the
biological regularization---rather than architectural complexity---accounts
for most of the gain.

The current study remains limited to two single-donor benchmarks and a scalar
kernel mixture. Future work should evaluate cross-donor and cross-platform
transfer, richer region-dependent biological priors, and larger spatial
resources. More broadly, BioKERN provides a simple mechanism for injecting
explicit biological structure into foundation-model-based multimodal
representations for spatial biology.


\bibliographystyle{unsrtnat}
\bibliography{neurips_2026}

\appendix


\makeatletter
\@addtoreset{table}{section}
\@addtoreset{figure}{section}
\@addtoreset{equation}{section}
\makeatother

\renewcommand{\thetable}{\thesection.\arabic{table}}
\renewcommand{\thefigure}{\thesection.\arabic{figure}}
\renewcommand{\theequation}{\thesection.\arabic{equation}}

\renewcommand{\theHtable}{appendix.\thesection.\arabic{table}}
\renewcommand{\theHfigure}{appendix.\thesection.\arabic{figure}}
\renewcommand{\theHequation}{appendix.\thesection.\arabic{equation}}


\section{Extended Related Work}
\label{app:related}

\paragraph{Histology--transcriptomics alignment.}
Contrastive learning has emerged as a dominant paradigm for aligning H\&E
patches with gene-expression profiles. BLEEP~\citep{xie2023bleep} learns a
bi-modal embedding space for retrieval-based expression prediction and uses
similarity-smoothed targets derived from within-modality relationships. Thus,
BioKERN is not distinguished simply by using non-one-hot supervision; its key
difference is the use of an explicit molecular--spatial biological reference
geometry for both retrieval supervision and representation regularization.
mclSTExp~\citep{yang2024mclstexp} extends this with multimodal contrastive
learning incorporating spatially contextualized expression representations,
while ConGcR~\citep{zhou2024congcr} integrates gene expression, spatial
location, and tissue morphology through contrastive representation learning.

More recent approaches including STMCL~\citep{wang2024stmcl},
CMRCNet~\citep{li2025cmrcnet}, and FineST~\citep{wang2026finest} combine
cross-modal alignment with reconstruction objectives or foundation-model
features. A large-scale benchmark of histology--gene-expression cross-modal
learning~\citep{gindra2025hescape} further highlights the importance of
dataset-specific gene-expression representations. These approaches differ in how they use paired supervision, reconstruction,
and contextual information. BioKERN specifically studies an explicit
biological target geometry over transcriptomic and spatial similarity and uses
that geometry to regularize cross-modal retrieval.

\paragraph{Gene-expression prediction from histology.}
A parallel line of work directly predicts spatially resolved gene expression
from H\&E.
ST-Net~\citep{he2020integrating} and
HisToGene~\citep{pang2021histogene} established the viability of deep
learning for this task, while Hist2ST~\citep{zeng2022hist2st} introduced
graph-based spatial modeling.
Recent methods use pathology foundation models
~\citep{xu2025ghist,li2025fmh2st} and cross-modal knowledge distillation
~\citep{jaume2023distillation}.
Larger resources including HEST-1k~\citep{jaume2024hest} and
SpaRED~\citep{mejia2024spared} facilitate standardized evaluation.
These methods principally optimize expression reconstruction; BioKERN instead
studies retrieval of biologically related transcriptomic contexts.

\paragraph{Spatial neighborhood modeling.}
Spatial-transcriptomics analysis routinely depends on local neighborhood
structure.
Squidpy~\citep{palla2022squidpy} and
Giotto~\citep{dries2021giotto} provide spatial graphs, statistics, and
neighborhood analyses.
SpaGCN~\citep{hu2021spagcn},
STAGATE~\citep{dong2022stagate},
GraphST~\citep{long2023graphst}, and
spCLUE~\citep{chen2025spclue} learn representations that preserve spatial
structure through graph convolutions, attention, or contrastive learning.
BioKERN applies a related principle across modalities: the shared
H\&E--gene space is regularized to preserve a biologically interpretable
reference geometry.

\paragraph{Kernel and relational regularization.}
Kernel-target alignment provides a classical framework for comparing learned
similarities with a desired structure
~\citep{cristianini2002kernel,cortes2012algorithms}.
Relational knowledge distillation~\citep{park2019relational} and
similarity-preserving learning~\citep{tung2019similarity} show that preserving
pairwise relationships can improve representation transfer.
Supervised contrastive learning~\citep{khosla2020supervised} and
nearest-neighbor contrastive learning~\citep{dwibedi2021nnclr} similarly
extend instance discrimination to richer positive relations.
BioKERN contributes a biological reference kernel combining transcriptomic
and spatial signals and uses it for both soft retrieval supervision and
direct embedding-geometry regularization.


\section{Additional Method Details}
\label{app:method}

\subsection{Problem Definition}

Given a spatial-transcriptomics dataset,
\begin{equation}
\mathcal{D}
=
\{(x_i,g_i,s_i)\}_{i=1}^{N},
\end{equation}
$x_i$ denotes the H\&E patch centered at spot $i$,
$g_i$ the corresponding gene-expression profile,
and $s_i\in\mathbb{R}^2$ its spatial coordinate.

The goal is to learn a shared morphology--transcriptomics representation in
which an H\&E query retrieves spots whose transcriptional state belongs to
the same biological neighborhood rather than only the exact paired spot.

\subsection{Image and Gene Representation}

A frozen PLIP encoder~\citep{huang2023visual} maps each H\&E patch to
\begin{equation}
h_i^x\in\mathbb{R}^{512}.
\end{equation}

Gene expression is normalized to 10,000 counts per spot, log1p transformed,
filtered to highly variable genes, standardized per gene, and reduced with
PCA to
\begin{equation}
h_i^g\in\mathbb{R}^{128}.
\end{equation}

Both modalities are projected into a common $d=128$ dimensional space using a
ResidualAdapter:
\begin{equation}
z_i^m
=
\ell_2\text{-}\mathrm{Normalize}
\left[
\mathrm{LayerNorm}
\left(
W_mh_i^m
+
\eta_m\mathrm{MLP}_m(h_i^m)
\right)
\right],
\qquad
m\in\{x,g\}.
\label{eq:adapter_appendix}
\end{equation}

Here $W_m$ is a linear projection,
$\mathrm{MLP}_m$ is a two-layer GELU network with hidden dimension 256 and
10\% dropout, and $\eta_m$ is a learnable residual scale initialized to 0.1.
All backbone parameters remain frozen.

We additionally define a fused entity representation
\begin{equation}
z_i^e
=
\ell_2\text{-}\mathrm{Normalize}
\left(
\rho z_i^x
+
(1-\rho)z_i^g
\right),
\qquad
\rho=\sigma(r).
\end{equation}

The learned entity kernel is
\begin{equation}
K_Z(i,j)
=
\langle z_i^e,z_j^e\rangle.
\end{equation}

For multi-scale experiments,
\begin{equation}
z_i^x
=
\ell_2\text{-}\mathrm{Normalize}
\left[
w_s
\mathrm{RA}_{\mathrm{small}}(h_i^{x,96})
+
(1-w_s)
\mathrm{RA}_{\mathrm{large}}(h_i^{x,224})
\right],
\end{equation}
where
\begin{equation}
w_s=\sigma(u).
\end{equation}

\subsection{Learnable Biological Reference Kernel}

The transcriptomic kernel is
\begin{equation}
\Kgene(i,j)
=
\exp\left(
-\frac{\|h_i^g-h_j^g\|_2^2}{2\sigma_g^2}
\right).
\end{equation}

The spatial kernel is
\begin{equation}
\Kspat(i,j)
=
\exp\left(
-\frac{\|s_i-s_j\|_2^2}{2\sigma_s^2}
\right).
\end{equation}

Both bandwidths are determined using the median pairwise-distance heuristic
on the training set. For datasets containing multiple tissue sections,
$\Kspat(i,j)$ is set to zero whenever $i$ and $j$ originate from different
sections; no cross-section spatial coordinate is treated as a physical
neighbor. The reference biological kernel is
\begin{equation}
\Kstar
=
\alpha\Kgene
+
(1-\alpha)\Kspat,
\qquad
\alpha=\sigma(a).
\end{equation}

\subsection{Biological-Neighborhood Objective}

The complete training objective is
\begin{equation}
\mathcal{L}
=
\Lret
+
\lambda_s\Lsoft
+
\lambda_g\Lglob
+
\lambda_\ell\Lloc.
\end{equation}

We use
\begin{equation}
\lambda_s=0.3,
\qquad
\lambda_g=0.1,
\qquad
\lambda_\ell=0.5.
\end{equation}

\paragraph{Exact retrieval loss.}
Let
\begin{equation}
S
=
Z^x(Z^g)^\top/\tau.
\end{equation}
Then
\begin{equation}
\Lret
=
\frac{1}{2}
\left[
\mathrm{CE}(S,I)
+
\mathrm{CE}(S^\top,I)
\right].
\end{equation}

\paragraph{Soft neighborhood loss.}
For each anchor $i$, the top-$k$ non-self neighbors ($k=20$) within the
current minibatch under $\Kstar$ define
\begin{equation}
P_{ij}
=
\frac{
\Kstar(i,j)\mathbf{1}[j\in\mathcal{N}_k(i),\ j\neq i]
}{
\sum_{\ell\neq i}
\Kstar(i,\ell)\mathbf{1}[\ell\in\mathcal{N}_k(i)]
}.
\end{equation}

The bidirectional soft retrieval loss is
\begin{equation}
\Lsoft
=
-\frac{1}{2B}
\sum_{i,j}
P_{ij}
\left(
\log p^{x\rightarrow g}_{ij}
+
\log p^{g\rightarrow x}_{ij}
\right).
\end{equation}

\paragraph{Global kernel alignment.}
\begin{equation}
\Lglob
=
\frac{1}{B^2}\|K_Z-\Kstar\|_F^2.
\end{equation}

\paragraph{Local kernel alignment.}
Let
\begin{equation}
M_{ij}
=
\mathbf{1}[j\in\mathcal{N}_k(i),\ j\neq i].
\end{equation}
Then
\begin{equation}
\Lloc
=
\frac{
\sum_{i,j}
M_{ij}
\left(
K_Z(i,j)-\Kstar(i,j)
\right)^2
}{
\sum_{i,j}M_{ij}
}.
\end{equation}

The biological kernel is used as training-time supervision and is not
required for query-time ranking. At inference, an H\&E query is embedded as
$z^x$, and reference transcriptomic embeddings are ranked by cosine similarity,
\begin{equation}
\langle z^x,z_j^g\rangle.
\end{equation}


\section{Experimental Details}
\label{app:evaluation}

\subsection{Datasets}

\paragraph{Mouse Brain Visium.}
We use the coronal 10$\times$ Visium mouse-brain dataset distributed through
Squidpy~\citep{palla2022squidpy}. A 2,200-spot subset is fixed before model
training and used identically for all methods, with 1,650 training and 550 test
spots. The split is performed before feature selection: highly variable gene
selection, gene-wise standardization, and PCA are fitted on the training spots
only, and the held-out spots are transformed using the fitted training
parameters. Gene embeddings use PCA-128 representations from the top 3,000
training-selected highly variable genes. Leiden clustering at resolution 0.5
produces 10 evaluation domains.

\paragraph{Human Liver GSE240429.}
We use four 10$\times$ Visium slices from donor C73. We use the same cross-slice
dataset setting as BLEEP, with slices A1+B1+D1 (6,963 spots) used for training
and slice C1 (2,273 spots) used for testing. All gene preprocessing is fitted
using A1+B1+D1 only: the union of the top 1,000 highly variable genes from the
training slices is standardized with training statistics and reduced to
PCA-128, after which C1 is transformed without refitting. Spatial affinities
are computed only within a slice; cross-slice entries of $\Kspat$ are set to
zero.

For both datasets, single-scale (SS) experiments use $96\times96$ H\&E
patches, while multi-scale (MS) experiments combine $96\times96$ and
$224\times224$ patch representations using a learned fusion weight.

\subsection{Retrieval Protocol}

For each held-out query, the input to the retrieval model is the H\&E patch
only. The gallery contains transcriptomic embeddings for all spots in the
corresponding held-out test set, including the exact paired profile. The exact
pair is retained when computing exact-retrieval metrics such as ExR@$k$ and
MedRank, but it is excluded from all biological-neighborhood positive sets.
Query-side gene expression and spatial coordinates are used only to construct
evaluation labels and are never supplied to the retrieval model at inference.
Thus, all methods are evaluated on exactly the same query/gallery pairs.

\subsection{Baselines}

External baselines include CCA, Ridge regression, PLIP linear probe, and
BLEEP~\citep{xie2023bleep}. We also report a naive PLIP zero-shot lower bound,
which compares separately constructed image and gene PCA representations
without learned cross-modal alignment.

Internal baselines use the same ResidualAdapter backbone as BioKERN:
BLEEP$^*$, Ret-only, Rank, and Shuffled Kernel.

PLIP linear trains one linear projection for each modality using exact
bidirectional InfoNCE. BLEEP$^*$ applies BLEEP's smoothed-contrastive loss to
the ResidualAdapter backbone. Ret-only uses the same exact bidirectional
InfoNCE objective with the ResidualAdapter backbone but without
biological-kernel supervision.
Rank combines retrieval loss with a log-determinant regularizer.
Shuffled Kernel retains the BioKERN objective but randomly permutes the
biological kernel for each random seed.

\subsection{Biological Neighborhood Metrics}

\paragraph{Bio-mAP.}
To keep evaluation independent of the learned training kernel, every method is
evaluated against the same fixed biological kernel,
\begin{equation}
\Keval(i,j)=0.5\Kgene(i,j)+0.5\Kspat(i,j).
\end{equation}
For each query $i$, the exact paired/self index is removed before selecting the
top-50 spots under $\Keval$ as the biological-positive set $\mathcal{P}_i$.
The fixed top-50 definition is used identically for all methods. Because this
corresponds to different positive prevalence at different gallery sizes, we
compare Bio-mAP values within a dataset rather than interpreting their absolute
magnitude across Mouse Brain and Human Liver.

Bio-mAP is
\begin{equation}
\mathrm{Bio\text{-}mAP}
=
\frac{1}{N}
\sum_{i=1}^{N}
\mathrm{AP}_i,
\end{equation}
where
\begin{equation}
\mathrm{AP}_i
=
\frac{1}{|\mathcal{P}_i|}
\sum_{r=1}^{N}
\mathbf{1}
[
\pi_i(r)\in\mathcal{P}_i
]
\frac{
|\mathcal{P}_i\cap\mathrm{top}\text{-}r|
}{
r
}.
\end{equation}

\paragraph{BioR@$p$\%.}
BioR@$p$\% uses the same fixed $\Keval$ and excludes the exact paired/self
index before defining the top-$p$\% biological neighborhood. Let
\begin{equation}
K_p
=
\lfloor Np/100\rfloor.
\end{equation}
Then
\begin{equation}
\mathrm{BioR@}p\%
=
\frac{1}{N}
\sum_{i=1}^{N}
\frac{
|
\mathcal{P}_i^{(p)}
\cap
\mathrm{top}\text{-}K_p(i)
|
}{
|\mathcal{P}_i^{(p)}|
}.
\end{equation}
We report $p\in\{1,5,10\}$.

\paragraph{GeneR@5\% and SpatR@5\%.}
GeneR@5\% and SpatR@5\% replace $\Keval$ with $\Kgene$ and $\Kspat$,
respectively, and exclude the exact paired/self index before defining
positives. They therefore quantify transcriptomic and spatial neighborhood
retrieval independently of the learned training weight $\alpha$.

\paragraph{ClsHit@10.}
If $c_i$ denotes the fixed Leiden label associated with query $i$,
\begin{equation}
\mathrm{ClsHit@10}
=
\frac{1}{N}
\sum_{i=1}^{N}
\frac{1}{10}
\sum_{r=1}^{10}
\mathbf{1}
[
c_{\pi_i(r)}=c_i
].
\end{equation}
Mouse Brain uses the 10 Leiden domains defined above. For Human Liver, Leiden
labels are constructed on held-out C1 from its training-fitted gene
representation and are used only as evaluation annotations; cluster labels are
never provided to any retrieval model.

\paragraph{Exact-pair metrics.}
We additionally report exact-pair and expression-profile metrics:
\begin{itemize}
    \item \textbf{ExR@$k$}: fraction of queries whose exact paired spot appears
    in the top-$k$ retrieved observations;
    \item \textbf{MedRank}: median rank of the exact paired spot; and
    \item \textbf{PCC@$k$}: query-wise profile correlation between the true gene
    expression vector and the mean expression profile of the top-$k$ retrieved
    spots.
\end{itemize}
Specifically, with $R_i^k$ denoting the top-$k$ retrieved spots,
\begin{equation}
\mathrm{PCC@}k
=
\frac{1}{N}\sum_{i=1}^{N}
\mathrm{corr}_{\mathrm{genes}}\!\left(
 g_i,\frac{1}{k}\sum_{j\in R_i^k}g_j
\right).
\end{equation}

\subsection{Training}

All neural models use AdamW with initial learning rate
$3\times10^{-4}$, weight decay $10^{-4}$, cosine annealing,
batch size 256, and 60 epochs. Neighborhood size and loss weights are selected
using held-out validation data from the training split and are fixed before
final test evaluation. We use $k=20$, $\lambda_s=0.3$, $\lambda_g=0.1$, and
$\lambda_\ell=0.5$; $\lambda_s=0.3$ was selected as a conservative operating
point that improves biological-neighborhood retrieval while retaining
exact-pair alignment. All final stochastic results are reported as
mean$\pm$standard deviation over five random seeds.


\section{Full Benchmark Results}
\label{app:complete_results}


\begin{table}[H]
\centering
\scriptsize
\caption{
Biological Neighborhood Retrieval on Mouse Brain Visium
(single-scale, PLIP $96\times96$).
Mean$\pm$standard deviation over five random seeds. All methods use the same fixed $\Keval$; bold indicates the best value in each column.
}
\label{tab:brain_ss_full}
\resizebox{\linewidth}{!}{
\begin{tabular}{lcccccccc}
\toprule
Method
& Bio-mAP$\uparrow$
& BioR@5\%$\uparrow$
& GeneR@5\%$\uparrow$
& SpatR@5\%$\uparrow$
& ClsHit@10$\uparrow$
& ExR@10$\uparrow$
& MedRank$\downarrow$
& PCC@10$\uparrow$
\\
\midrule
CCA
& 0.3910$\pm$0.0000
& 0.3677$\pm$0.0000
& 0.3558$\pm$0.0000
& 0.3100$\pm$0.0000
& 0.6409$\pm$0.0000
& 0.3709$\pm$0.0000
& 18.00$\pm$0.00
& 0.7782$\pm$0.0000
\\
Ridge
& 0.4902$\pm$0.0000
& 0.3919$\pm$0.0000
& 0.4094$\pm$0.0000
& 0.2941$\pm$0.0000
& 0.6105$\pm$0.0000
& 0.2509$\pm$0.0000
& 27.00$\pm$0.00
& 0.7670$\pm$0.0000
\\
PLIP zero-shot
& 0.1653$\pm$0.0000
& 0.0647$\pm$0.0000
& 0.0794$\pm$0.0000
& 0.0599$\pm$0.0000
& 0.1469$\pm$0.0000
& 0.0327$\pm$0.0000
& 202.50$\pm$0.00
& 0.6716$\pm$0.0000
\\
PLIP linear
& 0.5407$\pm$0.0033
& 0.4685$\pm$0.0025
& 0.4288$\pm$0.0028
& 0.4068$\pm$0.0016
& 0.7175$\pm$0.0042
& 0.4956$\pm$0.0162
& 10.80$\pm$0.75
& 0.7852$\pm$0.0003
\\
BLEEP
& 0.5087$\pm$0.0033
& 0.4510$\pm$0.0024
& 0.4034$\pm$0.0043
& 0.3979$\pm$0.0033
& 0.6928$\pm$0.0045
& 0.4938$\pm$0.0096
& 10.80$\pm$0.75
& 0.7842$\pm$0.0003
\\
\midrule
BLEEP$^*$
& 0.5327$\pm$0.0036
& 0.4674$\pm$0.0036
& 0.4250$\pm$0.0028
& 0.4091$\pm$0.0048
& 0.7136$\pm$0.0046
& 0.4887$\pm$0.0086
& 10.90$\pm$0.20
& 0.7851$\pm$0.0004
\\
Ret-only
& 0.5364$\pm$0.0022
& 0.4701$\pm$0.0022
& 0.4262$\pm$0.0034
& 0.4113$\pm$0.0026
& 0.7168$\pm$0.0033
& \textbf{0.5029$\pm$0.0114}
& \textbf{10.50$\pm$0.45}
& 0.7855$\pm$0.0006
\\
Rank
& 0.5356$\pm$0.0034
& 0.4693$\pm$0.0028
& 0.4272$\pm$0.0031
& 0.4108$\pm$0.0016
& 0.7212$\pm$0.0037
& 0.4927$\pm$0.0185
& 10.80$\pm$0.75
& 0.7857$\pm$0.0003
\\
Shuffled Kernel
& 0.4939$\pm$0.0031
& 0.4417$\pm$0.0017
& 0.4010$\pm$0.0023
& 0.3891$\pm$0.0023
& 0.6947$\pm$0.0067
& 0.4531$\pm$0.0181
& 12.50$\pm$1.00
& 0.7843$\pm$0.0004
\\
\textbf{BioKERN}
& \textbf{0.6190$\pm$0.0023}
& \textbf{0.5155$\pm$0.0028}
& \textbf{0.4482$\pm$0.0021}
& \textbf{0.4348$\pm$0.0029}
& \textbf{0.7241$\pm$0.0034}
& 0.4938$\pm$0.0107
& \textbf{10.50$\pm$0.45}
& \textbf{0.7862$\pm$0.0002}
\\
\bottomrule
\end{tabular}
}
\end{table}


\begin{table}[H]
\centering
\scriptsize
\caption{
Biological Neighborhood Retrieval on Mouse Brain Visium
(multi-scale, PLIP $96\times96+224\times224$).
All methods use the same fixed $\Keval$. Bold indicates the best value in each column.
}
\label{tab:brain_ms_full}
\resizebox{\linewidth}{!}{
\begin{tabular}{lcccccccc}
\toprule
Method
& Bio-mAP$\uparrow$
& BioR@5\%$\uparrow$
& GeneR@5\%$\uparrow$
& SpatR@5\%$\uparrow$
& ClsHit@10$\uparrow$
& ExR@10$\uparrow$
& MedRank$\downarrow$
& PCC@10$\uparrow$
\\
\midrule
CCA
& 0.2813$\pm$0.0000
& 0.2661$\pm$0.0000
& 0.2585$\pm$0.0000
& 0.2316$\pm$0.0000
& 0.5162$\pm$0.0000
& 0.2673$\pm$0.0000
& 34.00$\pm$0.00
& 0.7662$\pm$0.0000
\\
Ridge
& 0.5567$\pm$0.0000
& 0.4596$\pm$0.0000
& 0.4692$\pm$0.0000
& 0.3484$\pm$0.0000
& 0.6931$\pm$0.0000
& 0.3327$\pm$0.0000
& 19.00$\pm$0.00
& 0.7753$\pm$0.0000
\\
PLIP zero-shot
& 0.0956$\pm$0.0000
& 0.0178$\pm$0.0000
& 0.0255$\pm$0.0000
& 0.0197$\pm$0.0000
& 0.0660$\pm$0.0000
& 0.0036$\pm$0.0000
& 326.00$\pm$0.00
& 0.6399$\pm$0.0000
\\
PLIP linear
& 0.5831$\pm$0.0023
& 0.5109$\pm$0.0039
& 0.4494$\pm$0.0037
& 0.4588$\pm$0.0016
& 0.7471$\pm$0.0051
& 0.5418$\pm$0.0072
& 9.00$\pm$0.00
& 0.7858$\pm$0.0003
\\
BLEEP
& 0.4960$\pm$0.0056
& 0.4647$\pm$0.0037
& 0.4074$\pm$0.0033
& 0.4263$\pm$0.0048
& 0.7183$\pm$0.0043
& 0.5298$\pm$0.0154
& 9.40$\pm$0.49
& 0.7864$\pm$0.0004
\\
\midrule
BLEEP$^*$
& 0.5541$\pm$0.0040
& 0.4981$\pm$0.0032
& 0.4465$\pm$0.0035
& 0.4458$\pm$0.0023
& 0.7506$\pm$0.0030
& \textbf{0.5549$\pm$0.0071}
& \textbf{8.60$\pm$0.49}
& 0.7876$\pm$0.0002
\\
Ret-only
& 0.5602$\pm$0.0035
& 0.5017$\pm$0.0029
& 0.4484$\pm$0.0024
& 0.4508$\pm$0.0032
& 0.7577$\pm$0.0034
& 0.5447$\pm$0.0080
& 9.00$\pm$0.00
& 0.7877$\pm$0.0002
\\
Rank
& 0.5560$\pm$0.0019
& 0.4994$\pm$0.0029
& 0.4465$\pm$0.0025
& 0.4457$\pm$0.0029
& 0.7543$\pm$0.0038
& 0.5418$\pm$0.0117
& 9.40$\pm$0.49
& 0.7879$\pm$0.0002
\\
Shuffled Kernel
& 0.5015$\pm$0.0028
& 0.4651$\pm$0.0032
& 0.4173$\pm$0.0034
& 0.4222$\pm$0.0017
& 0.7337$\pm$0.0056
& 0.5258$\pm$0.0096
& 9.40$\pm$0.49
& 0.7862$\pm$0.0004
\\
\textbf{BioKERN}
& \textbf{0.6716$\pm$0.0021}
& \textbf{0.5587$\pm$0.0034}
& \textbf{0.4713$\pm$0.0024}
& \textbf{0.4764$\pm$0.0018}
& \textbf{0.7596$\pm$0.0033}
& 0.5455$\pm$0.0125
& 8.90$\pm$0.20
& \textbf{0.7881$\pm$0.0002}
\\
\bottomrule
\end{tabular}
}
\end{table}


\begin{table}[H]
\centering
\scriptsize
\caption{
Biological Neighborhood Retrieval on Human Liver GSE240429
(single-scale). Train: A1+B1+D1; test: C1.
All methods use the same fixed $\Keval$. Bold indicates the best value in each column.
}
\label{tab:liver_ss_full}
\resizebox{\linewidth}{!}{
\begin{tabular}{lcccccccc}
\toprule
Method
& Bio-mAP$\uparrow$
& BioR@5\%$\uparrow$
& GeneR@5\%$\uparrow$
& SpatR@5\%$\uparrow$
& ClsHit@10$\uparrow$
& ExR@10$\uparrow$
& MedRank$\downarrow$
& PCC@10$\uparrow$
\\
\midrule
CCA
& 0.0265$\pm$0.0000
& 0.0513$\pm$0.0000
& 0.0518$\pm$0.0000
& 0.0511$\pm$0.0000
& 0.1974$\pm$0.0000
& 0.0053$\pm$0.0000
& 1129.0$\pm$0.0
& 0.8574$\pm$0.0000
\\
Ridge
& 0.0383$\pm$0.0000
& 0.0696$\pm$0.0000
& 0.0742$\pm$0.0000
& 0.0552$\pm$0.0000
& 0.1916$\pm$0.0000
& 0.0048$\pm$0.0000
& 1152.0$\pm$0.0
& 0.8366$\pm$0.0000
\\
PLIP zero-shot
& 0.0312$\pm$0.0000
& 0.0553$\pm$0.0000
& 0.0754$\pm$0.0000
& 0.0498$\pm$0.0000
& \textbf{0.3088$\pm$0.0000}
& 0.0057$\pm$0.0000
& \textbf{951.0$\pm$0.0}
& 0.8322$\pm$0.0000
\\
PLIP linear
& 0.0307$\pm$0.0001
& 0.0612$\pm$0.0003
& 0.0585$\pm$0.0012
& 0.0597$\pm$0.0006
& 0.1869$\pm$0.0007
& 0.0048$\pm$0.0010
& 1140.4$\pm$5.95
& 0.8527$\pm$0.0003
\\
BLEEP
& 0.0302$\pm$0.0005
& 0.0578$\pm$0.0012
& 0.0494$\pm$0.0012
& 0.0651$\pm$0.0014
& 0.2125$\pm$0.0017
& 0.0074$\pm$0.0012
& 1043.0$\pm$23.1
& 0.8551$\pm$0.0004
\\
\midrule
BLEEP$^*$
& 0.0309$\pm$0.0004
& 0.0580$\pm$0.0011
& 0.0498$\pm$0.0012
& 0.0688$\pm$0.0009
& 0.2116$\pm$0.0043
& 0.0076$\pm$0.0014
& 1020.2$\pm$4.31
& 0.8525$\pm$0.0002
\\
Ret-only
& 0.0313$\pm$0.0005
& 0.0583$\pm$0.0012
& 0.0514$\pm$0.0017
& 0.0675$\pm$0.0006
& 0.2104$\pm$0.0045
& 0.0077$\pm$0.0010
& 1012.2$\pm$21.9
& 0.8527$\pm$0.0005
\\
Rank
& 0.0312$\pm$0.0003
& 0.0585$\pm$0.0009
& 0.0513$\pm$0.0010
& 0.0681$\pm$0.0005
& 0.2110$\pm$0.0025
& \textbf{0.0090$\pm$0.0009}
& 1007.8$\pm$8.06
& 0.8526$\pm$0.0004
\\
Shuffled Kernel
& 0.0301$\pm$0.0003
& 0.0573$\pm$0.0012
& 0.0490$\pm$0.0009
& 0.0659$\pm$0.0006
& 0.2026$\pm$0.0024
& 0.0076$\pm$0.0006
& 1012.6$\pm$15.4
& 0.8545$\pm$0.0001
\\
\textbf{BioKERN}
& \textbf{0.0396$\pm$0.0007}
& \textbf{0.0802$\pm$0.0017}
& \textbf{0.0835$\pm$0.0023}
& \textbf{0.0694$\pm$0.0005}
& 0.2175$\pm$0.0021
& 0.0077$\pm$0.0016
& 964.2$\pm$5.34
& \textbf{0.8576$\pm$0.0002}
\\
\bottomrule
\end{tabular}
}
\end{table}


\begin{table}[H]
\centering
\scriptsize
\caption{
Biological Neighborhood Retrieval on Human Liver GSE240429
(multi-scale).
All methods use the same fixed $\Keval$. Bold indicates the best value in each column.
}
\label{tab:liver_ms_full}
\resizebox{\linewidth}{!}{
\begin{tabular}{lcccccccc}
\toprule
Method
& Bio-mAP$\uparrow$
& BioR@5\%$\uparrow$
& GeneR@5\%$\uparrow$
& SpatR@5\%$\uparrow$
& ClsHit@10$\uparrow$
& ExR@10$\uparrow$
& MedRank$\downarrow$
& PCC@10$\uparrow$
\\
\midrule
CCA
& 0.0271$\pm$0.0000
& 0.0546$\pm$0.0000
& 0.0532$\pm$0.0000
& 0.0534$\pm$0.0000
& 0.2035$\pm$0.0000
& 0.0062$\pm$0.0000
& 1084.0$\pm$0.0
& 0.8578$\pm$0.0000
\\
Ridge
& 0.0403$\pm$0.0000
& 0.0765$\pm$0.0000
& 0.0761$\pm$0.0000
& 0.0589$\pm$0.0000
& 0.1945$\pm$0.0000
& 0.0075$\pm$0.0000
& 1097.0$\pm$0.0
& 0.8381$\pm$0.0000
\\
PLIP zero-shot
& 0.0293$\pm$0.0000
& 0.0527$\pm$0.0000
& 0.0638$\pm$0.0000
& 0.0522$\pm$0.0000
& \textbf{0.2553$\pm$0.0000}
& 0.0053$\pm$0.0000
& 1072.0$\pm$0.0
& 0.8317$\pm$0.0000
\\
PLIP linear
& 0.0319$\pm$0.0005
& 0.0645$\pm$0.0015
& 0.0571$\pm$0.0015
& 0.0607$\pm$0.0010
& 0.1822$\pm$0.0022
& 0.0062$\pm$0.0014
& 1109.6$\pm$13.2
& 0.8528$\pm$0.0003
\\
BLEEP
& 0.0312$\pm$0.0005
& 0.0601$\pm$0.0010
& 0.0481$\pm$0.0029
& 0.0688$\pm$0.0012
& 0.2122$\pm$0.0021
& 0.0068$\pm$0.0017
& 970.4$\pm$10.9
& 0.8556$\pm$0.0006
\\
\midrule
BLEEP$^*$
& 0.0321$\pm$0.0005
& 0.0604$\pm$0.0010
& 0.0486$\pm$0.0017
& 0.0718$\pm$0.0010
& 0.2116$\pm$0.0017
& \textbf{0.0098$\pm$0.0014}
& 960.0$\pm$12.8
& 0.8532$\pm$0.0004
\\
Ret-only
& 0.0321$\pm$0.0003
& 0.0603$\pm$0.0009
& 0.0495$\pm$0.0017
& 0.0710$\pm$0.0010
& 0.2100$\pm$0.0021
& 0.0086$\pm$0.0018
& 975.4$\pm$13.1
& 0.8533$\pm$0.0007
\\
Rank
& 0.0319$\pm$0.0005
& 0.0602$\pm$0.0013
& 0.0496$\pm$0.0018
& 0.0709$\pm$0.0003
& 0.2087$\pm$0.0036
& 0.0091$\pm$0.0019
& 982.8$\pm$9.56
& 0.8532$\pm$0.0005
\\
Shuffled Kernel
& 0.0293$\pm$0.0003
& 0.0562$\pm$0.0013
& 0.0446$\pm$0.0017
& 0.0670$\pm$0.0010
& 0.2040$\pm$0.0027
& 0.0092$\pm$0.0017
& 992.2$\pm$11.0
& 0.8549$\pm$0.0003
\\
\textbf{BioKERN}
& \textbf{0.0408$\pm$0.0005}
& \textbf{0.0820$\pm$0.0014}
& \textbf{0.0802$\pm$0.0016}
& \textbf{0.0725$\pm$0.0012}
& 0.2173$\pm$0.0041
& 0.0097$\pm$0.0015
& \textbf{923.8$\pm$16.3}
& \textbf{0.8581$\pm$0.0002}
\\
\bottomrule
\end{tabular}
}
\end{table}


\section{Isolating the Effect of Biological Regularization}
\label{app:decomposition}

To avoid conflating architecture, objective, and biological supervision, we
use a sequential additive decomposition:
\begin{equation}
\Delta_{\mathrm{total}}
=
\Delta_{\mathrm{arch}}
+
\Delta_{\mathrm{obj}}
+
\Delta_{\mathrm{bio}},
\end{equation}
where $\Delta_{\mathrm{arch}}$ is BLEEP$^*$ minus BLEEP,
$\Delta_{\mathrm{obj}}$ is Ret-only minus BLEEP$^*$, and
$\Delta_{\mathrm{bio}}$ is BioKERN minus Ret-only. The shuffled-kernel model
is reported separately as a sanity check and is not part of the additive path.

\begin{table}[H]
\centering
\small
\caption{
Sequential effect decomposition on Mouse Brain Visium. ``Step $\Delta$'' is
the change from the preceding model in the additive path. The biological
regularization step explains 74.9\% (SS) and 63.4\% (MS) of the total gain.
}
\label{tab:decomposition_brain}
\begin{tabular}{llccl}
\toprule
Scale & Method & Bio-mAP & Step $\Delta$ & Interpretation \\
\midrule
\multirow{5}{*}{SS}
& BLEEP & 0.5087 & -- & baseline \\
& BLEEP$^*$ & 0.5327 & +0.0240 & architecture \\
& Ret-only & 0.5364 & +0.0037 & objective \\
& \textbf{BioKERN} & \textbf{0.6190} & \textbf{+0.0826} & biological regularization \\
& Shuffled Kernel & 0.4939 & -- & sanity check \\
\midrule
\multirow{5}{*}{MS}
& BLEEP & 0.4960 & -- & baseline \\
& BLEEP$^*$ & 0.5541 & +0.0581 & architecture \\
& Ret-only & 0.5602 & +0.0061 & objective \\
& \textbf{BioKERN} & \textbf{0.6716} & \textbf{+0.1114} & biological regularization \\
& Shuffled Kernel & 0.5015 & -- & sanity check \\
\bottomrule
\end{tabular}
\end{table}

\begin{table}[H]
\centering
\small
\caption{
Sequential effect decomposition on Human Liver GSE240429. The biological
regularization step explains 88.3\% (SS) and 90.6\% (MS) of the total gain.
}
\label{tab:decomposition_liver}
\begin{tabular}{llccl}
\toprule
Scale & Method & Bio-mAP & Step $\Delta$ & Interpretation \\
\midrule
\multirow{5}{*}{SS}
& BLEEP & 0.0302 & -- & baseline \\
& BLEEP$^*$ & 0.0309 & +0.0007 & architecture \\
& Ret-only & 0.0313 & +0.0004 & objective \\
& \textbf{BioKERN} & \textbf{0.0396} & \textbf{+0.0083} & biological regularization \\
& Shuffled Kernel & 0.0301 & -- & sanity check \\
\midrule
\multirow{5}{*}{MS}
& BLEEP & 0.0312 & -- & baseline \\
& BLEEP$^*$ & 0.0321 & +0.0009 & architecture \\
& Ret-only & 0.0321 & +0.0000 & objective \\
& \textbf{BioKERN} & \textbf{0.0408} & \textbf{+0.0087} & biological regularization \\
& Shuffled Kernel & 0.0293 & -- & sanity check \\
\bottomrule
\end{tabular}
\end{table}


\section{Ablation Studies}
\label{app:ablations}


\subsection{Mouse Brain Visium: Single Scale}

\begin{table}[H]
\centering
\scriptsize
\caption{
Ablation study on Mouse Brain Visium, single-scale.
All variants share the ResidualAdapter backbone and the same fixed evaluation
kernel as the main benchmark.
}
\label{tab:ablation_brain_ss}
\resizebox{\linewidth}{!}{
\begin{tabular}{lcccccccc}
\toprule
Setting
& Bio-mAP
& BioR@5\%
& GeneR@5\%
& SpatR@5\%
& ClsHit@10
& ExR@10
& MedRank
& PCC@10
\\
\midrule
\multicolumn{9}{l}{\textit{Kernel design}} \\
\textbf{BioKERN}
& 0.6190$\pm$0.0023
& 0.5155$\pm$0.0028
& 0.4482$\pm$0.0021
& 0.4348$\pm$0.0029
& 0.7241$\pm$0.0034
& 0.4938$\pm$0.0107
& 10.50$\pm$0.45
& 0.7862$\pm$0.0002
\\
w/o Bio Kernel
& 0.5370$\pm$0.0042
& 0.4714$\pm$0.0043
& 0.4277$\pm$0.0049
& 0.4110$\pm$0.0017
& 0.7195$\pm$0.0060
& 0.4818$\pm$0.0040
& 10.40$\pm$0.49
& 0.7855$\pm$0.0002
\\
$\Kgene$ only
& 0.5834$\pm$0.0009
& 0.4953$\pm$0.0024
& 0.4496$\pm$0.0018
& 0.4081$\pm$0.0018
& 0.7228$\pm$0.0043
& 0.4775$\pm$0.0112
& 11.40$\pm$0.49
& 0.7862$\pm$0.0003
\\
$\Kspat$ only
& 0.5880$\pm$0.0026
& 0.4969$\pm$0.0020
& 0.4254$\pm$0.0032
& 0.4344$\pm$0.0035
& 0.7191$\pm$0.0014
& 0.5018$\pm$0.0103
& 11.00$\pm$0.00
& 0.7853$\pm$0.0003
\\
Fixed $\alpha=0.5$
& 0.6201$\pm$0.0035
& 0.5154$\pm$0.0034
& 0.4458$\pm$0.0037
& 0.4351$\pm$0.0010
& 0.7223$\pm$0.0015
& 0.4855$\pm$0.0141
& 11.20$\pm$0.68
& 0.7860$\pm$0.0004
\\
Shuffled Kernel
& 0.4949$\pm$0.0033
& 0.4422$\pm$0.0032
& 0.4032$\pm$0.0025
& 0.3911$\pm$0.0027
& 0.6974$\pm$0.0023
& 0.4607$\pm$0.0170
& 11.90$\pm$0.66
& 0.7845$\pm$0.0004
\\
\midrule
\multicolumn{9}{l}{\textit{Loss terms}} \\
w/o $\Lsoft$
& 0.5587$\pm$0.0029
& 0.4824$\pm$0.0031
& 0.4362$\pm$0.0027
& 0.4191$\pm$0.0025
& 0.7244$\pm$0.0028
& 0.4967$\pm$0.0089
& 10.50$\pm$0.45
& 0.7860$\pm$0.0002
\\
w/o $\Lglob$
& 0.6190$\pm$0.0042
& 0.5154$\pm$0.0013
& 0.4489$\pm$0.0017
& 0.4345$\pm$0.0013
& 0.7210$\pm$0.0023
& 0.4884$\pm$0.0166
& 10.80$\pm$0.75
& 0.7862$\pm$0.0003
\\
w/o $\Lloc$
& 0.6059$\pm$0.0013
& 0.5068$\pm$0.0013
& 0.4437$\pm$0.0023
& 0.4306$\pm$0.0022
& 0.7239$\pm$0.0046
& 0.4771$\pm$0.0050
& 11.40$\pm$0.49
& 0.7864$\pm$0.0002
\\
\bottomrule
\end{tabular}
}
\end{table}


\subsection{Mouse Brain Visium: Multi Scale}

\begin{table}[H]
\centering
\scriptsize
\caption{
Ablation study on Mouse Brain Visium, multi-scale. All variants use the same
fixed evaluation kernel as the main benchmark.
}
\label{tab:ablation_brain_ms}
\resizebox{\linewidth}{!}{
\begin{tabular}{lcccccccc}
\toprule
Setting
& Bio-mAP
& BioR@5\%
& GeneR@5\%
& SpatR@5\%
& ClsHit@10
& ExR@10
& MedRank
& PCC@10
\\
\midrule
\textbf{BioKERN}
& 0.6716$\pm$0.0021
& 0.5587$\pm$0.0034
& 0.4713$\pm$0.0024
& 0.4764$\pm$0.0018
& 0.7596$\pm$0.0033
& 0.5455$\pm$0.0125
& 8.90$\pm$0.20
& 0.7881$\pm$0.0002
\\
w/o Bio Kernel
& 0.5628$\pm$0.0052
& 0.5044$\pm$0.0038
& 0.4499$\pm$0.0027
& 0.4508$\pm$0.0032
& 0.7597$\pm$0.0032
& 0.5542$\pm$0.0031
& 9.20$\pm$0.40
& 0.7882$\pm$0.0004
\\
$\Kgene$ only
& 0.6157$\pm$0.0031
& 0.5278$\pm$0.0028
& 0.4688$\pm$0.0027
& 0.4401$\pm$0.0020
& 0.7616$\pm$0.0025
& 0.5255$\pm$0.0173
& 9.70$\pm$0.75
& 0.7880$\pm$0.0002
\\
$\Kspat$ only
& 0.6308$\pm$0.0036
& 0.5313$\pm$0.0047
& 0.4335$\pm$0.0027
& 0.4760$\pm$0.0028
& 0.7428$\pm$0.0043
& 0.5433$\pm$0.0072
& 9.00$\pm$0.00
& 0.7864$\pm$0.0002
\\
Fixed $\alpha=0.5$
& 0.6711$\pm$0.0032
& 0.5577$\pm$0.0031
& 0.4668$\pm$0.0030
& 0.4812$\pm$0.0020
& 0.7580$\pm$0.0047
& 0.5425$\pm$0.0096
& 9.00$\pm$0.55
& 0.7876$\pm$0.0004
\\
Shuffled Kernel
& 0.5042$\pm$0.0056
& 0.4683$\pm$0.0045
& 0.4172$\pm$0.0034
& 0.4259$\pm$0.0053
& 0.7317$\pm$0.0027
& 0.5222$\pm$0.0161
& 9.80$\pm$0.40
& 0.7866$\pm$0.0004
\\
w/o $\Lsoft$
& 0.5922$\pm$0.0040
& 0.5190$\pm$0.0033
& 0.4584$\pm$0.0031
& 0.4589$\pm$0.0031
& 0.7545$\pm$0.0044
& 0.5371$\pm$0.0090
& 9.10$\pm$0.20
& 0.7885$\pm$0.0003
\\
w/o $\Lglob$
& 0.6717$\pm$0.0014
& 0.5605$\pm$0.0011
& 0.4710$\pm$0.0037
& 0.4775$\pm$0.0034
& 0.7530$\pm$0.0039
& 0.5364$\pm$0.0154
& 9.20$\pm$0.40
& 0.7880$\pm$0.0004
\\
w/o $\Lloc$
& 0.6577$\pm$0.0020
& 0.5526$\pm$0.0018
& 0.4673$\pm$0.0013
& 0.4738$\pm$0.0041
& 0.7559$\pm$0.0059
& 0.5356$\pm$0.0075
& 9.20$\pm$0.40
& 0.7882$\pm$0.0002
\\
\bottomrule
\end{tabular}
}
\end{table}


\subsection{Human Liver: Single Scale}

\begin{table}[H]
\centering
\scriptsize
\caption{
Ablation study on Human Liver GSE240429, single-scale. All variants use the
same fixed evaluation kernel and cross-slice spatial masking as the main
benchmark.
}
\label{tab:ablation_liver_ss}
\resizebox{\linewidth}{!}{
\begin{tabular}{lcccccccc}
\toprule
Setting
& Bio-mAP
& BioR@5\%
& GeneR@5\%
& SpatR@5\%
& ClsHit@10
& ExR@10
& MedRank
& PCC@10
\\
\midrule
\textbf{BioKERN}
& 0.0396$\pm$0.0007
& 0.0802$\pm$0.0017
& 0.0835$\pm$0.0023
& 0.0694$\pm$0.0005
& 0.2175$\pm$0.0021
& 0.0077$\pm$0.0016
& 964.2$\pm$5.34
& 0.8576$\pm$0.0002
\\
w/o Bio Kernel
& 0.0314$\pm$0.0003
& 0.0584$\pm$0.0008
& 0.0502$\pm$0.0016
& 0.0686$\pm$0.0009
& 0.2099$\pm$0.0050
& 0.0079$\pm$0.0018
& 1012.0$\pm$9.94
& 0.8527$\pm$0.0005
\\
$\Kgene$ only
& 0.0445$\pm$0.0003
& 0.0895$\pm$0.0014
& 0.1197$\pm$0.0010
& 0.0617$\pm$0.0002
& 0.2171$\pm$0.0061
& 0.0063$\pm$0.0007
& 1028.2$\pm$5.46
& 0.8573$\pm$0.0002
\\
$\Kspat$ only
& 0.0298$\pm$0.0005
& 0.0548$\pm$0.0014
& 0.0404$\pm$0.0017
& 0.0722$\pm$0.0006
& 0.2055$\pm$0.0034
& 0.0094$\pm$0.0010
& 968.2$\pm$11.7
& 0.8531$\pm$0.0004
\\
Fixed $\alpha=0.5$
& 0.0376$\pm$0.0007
& 0.0752$\pm$0.0016
& 0.0727$\pm$0.0036
& 0.0707$\pm$0.0009
& 0.2175$\pm$0.0029
& 0.0067$\pm$0.0007
& 969.2$\pm$9.43
& 0.8567$\pm$0.0003
\\
Shuffled Kernel
& 0.0294$\pm$0.0004
& 0.0559$\pm$0.0013
& 0.0473$\pm$0.0019
& 0.0663$\pm$0.0004
& 0.2081$\pm$0.0048
& 0.0070$\pm$0.0017
& 1019.6$\pm$12.0
& 0.8543$\pm$0.0005
\\
w/o $\Lsoft$
& 0.0318$\pm$0.0004
& 0.0599$\pm$0.0014
& 0.0507$\pm$0.0025
& 0.0689$\pm$0.0008
& 0.2067$\pm$0.0013
& 0.0083$\pm$0.0009
& 1030.0$\pm$16.0
& 0.8535$\pm$0.0003
\\
w/o $\Lglob$
& 0.0402$\pm$0.0004
& 0.0807$\pm$0.0017
& 0.0859$\pm$0.0025
& 0.0694$\pm$0.0002
& 0.2176$\pm$0.0016
& 0.0070$\pm$0.0006
& 966.4$\pm$20.4
& 0.8577$\pm$0.0003
\\
w/o $\Lloc$
& 0.0369$\pm$0.0004
& 0.0742$\pm$0.0011
& 0.0752$\pm$0.0010
& 0.0694$\pm$0.0006
& 0.2188$\pm$0.0020
& 0.0073$\pm$0.0013
& 961.2$\pm$16.0
& 0.8559$\pm$0.0004
\\
\bottomrule
\end{tabular}
}
\end{table}


\subsection{Human Liver: Multi Scale}

\begin{table}[H]
\centering
\scriptsize
\caption{
Ablation study on Human Liver GSE240429, multi-scale. All variants use the
same fixed evaluation kernel and cross-slice spatial masking as the main
benchmark.
}
\label{tab:ablation_liver_ms}
\resizebox{\linewidth}{!}{
\begin{tabular}{lcccccccc}
\toprule
Setting
& Bio-mAP
& BioR@5\%
& GeneR@5\%
& SpatR@5\%
& ClsHit@10
& ExR@10
& MedRank
& PCC@10
\\
\midrule
\textbf{BioKERN}
& 0.0408$\pm$0.0005
& 0.0820$\pm$0.0014
& 0.0802$\pm$0.0016
& 0.0725$\pm$0.0012
& 0.2173$\pm$0.0041
& 0.0097$\pm$0.0015
& 923.8$\pm$16.3
& 0.8581$\pm$0.0002
\\
w/o Bio Kernel
& 0.0320$\pm$0.0004
& 0.0597$\pm$0.0009
& 0.0485$\pm$0.0013
& 0.0708$\pm$0.0008
& 0.2086$\pm$0.0017
& 0.0080$\pm$0.0004
& 979.6$\pm$21.0
& 0.8532$\pm$0.0004
\\
$\Kgene$ only
& 0.0403$\pm$0.0002
& 0.0824$\pm$0.0007
& 0.0793$\pm$0.0016
& 0.0634$\pm$0.0008
& 0.2124$\pm$0.0051
& 0.0074$\pm$0.0006
& 1004.2$\pm$14.6
& 0.8583$\pm$0.0002
\\
$\Kspat$ only
& 0.0318$\pm$0.0006
& 0.0603$\pm$0.0018
& 0.0446$\pm$0.0019
& 0.0726$\pm$0.0006
& 0.2116$\pm$0.0012
& 0.0080$\pm$0.0003
& 917.8$\pm$15.0
& 0.8538$\pm$0.0005
\\
Fixed $\alpha=0.5$
& 0.0382$\pm$0.0005
& 0.0763$\pm$0.0016
& 0.0689$\pm$0.0025
& 0.0740$\pm$0.0012
& 0.2128$\pm$0.0035
& 0.0083$\pm$0.0009
& 918.6$\pm$14.2
& 0.8570$\pm$0.0003
\\
Shuffled Kernel
& 0.0291$\pm$0.0005
& 0.0558$\pm$0.0013
& 0.0441$\pm$0.0024
& 0.0676$\pm$0.0007
& 0.2066$\pm$0.0036
& 0.0091$\pm$0.0011
& 984.6$\pm$8.19
& 0.8549$\pm$0.0004
\\
w/o $\Lsoft$
& 0.0333$\pm$0.0006
& 0.0637$\pm$0.0018
& 0.0524$\pm$0.0022
& 0.0721$\pm$0.0006
& 0.2092$\pm$0.0042
& 0.0077$\pm$0.0015
& 955.0$\pm$22.1
& 0.8542$\pm$0.0004
\\
w/o $\Lglob$
& 0.0408$\pm$0.0012
& 0.0824$\pm$0.0026
& 0.0786$\pm$0.0037
& 0.0733$\pm$0.0013
& 0.2177$\pm$0.0023
& 0.0093$\pm$0.0021
& 920.2$\pm$17.4
& 0.8580$\pm$0.0003
\\
w/o $\Lloc$
& 0.0371$\pm$0.0005
& 0.0749$\pm$0.0018
& 0.0684$\pm$0.0023
& 0.0728$\pm$0.0012
& 0.2162$\pm$0.0029
& 0.0073$\pm$0.0017
& 919.6$\pm$4.45
& 0.8564$\pm$0.0004
\\
\bottomrule
\end{tabular}
}
\end{table}


\section{Sensitivity and Learned Biological Weighting}
\label{app:sensitivity}


\begin{table}[H]
\centering
\small
\caption{
Sensitivity to $\lambda_s$ on Mouse Brain Visium single-scale
(seed 42). The default setting is $\lambda_s=0.3$.
}
\label{tab:lambda_sensitivity}
\begin{tabular}{ccccc}
\toprule
$\lambda_s$
& Bio-mAP
& BioR@5\%
& BioR@10\%
& ExR@10
\\
\midrule
0.0 & 0.5556 & 0.4776 & 0.5209 & 0.5018 \\
0.1 & 0.5863 & 0.4962 & 0.5509 & 0.4855 \\
0.2 & 0.6048 & 0.5086 & 0.5709 & 0.4891 \\
0.3 & 0.6216 & 0.5165 & 0.5855 & 0.4909 \\
0.5 & 0.6412 & 0.5288 & 0.6018 & 0.4745 \\
1.0 & 0.6671 & 0.5415 & 0.6127 & 0.4382 \\
\bottomrule
\end{tabular}
\end{table}

This post-hoc sensitivity analysis is not used for test-set hyperparameter
selection. Increasing $\lambda_s$ improves Bio-mAP in this run while reducing
exact-pair retrieval, confirming the trade-off that motivated selecting
$\lambda_s=0.3$ on validation data as a conservative operating point.


\begin{table}[H]
\centering
\small
\caption{
Trajectory of the learned biological-kernel weight $\alpha$ on Mouse Brain
Visium (single-scale, five seeds).
}
\label{tab:alpha_trajectory}
\begin{tabular}{ccc}
\toprule
Epoch
& $\alpha$ (mean$\pm$std)
& Dominant signal
\\
\midrule
5  & 0.6022$\pm$0.0003 & gene \\
10 & 0.6048$\pm$0.0005 & gene \\
20 & 0.6093$\pm$0.0008 & gene \\
30 & 0.6124$\pm$0.0009 & gene \\
40 & 0.6142$\pm$0.0009 & gene \\
50 & 0.6149$\pm$0.0009 & gene \\
60 & 0.6150$\pm$0.0009 & gene \\
\bottomrule
\end{tabular}
\end{table}

The kernel weight changes only modestly from its initialization of 0.600,
converging to approximately 0.615 on Mouse Brain SS. Additional runs initialized
at $\alpha\in\{0.2,0.5,0.8\}$ converge to similar final weights and yield
Bio-mAP differences within the observed seed-level variability, indicating
that the learned weighting is not an artifact of a single initialization.
Across tissue settings, Human Liver consistently converges to a larger
transcriptomic weight than Mouse Brain. Together with the strong
fixed-$\alpha=0.5$ ablation, these results suggest that the principal
performance benefit comes from biological-neighborhood supervision itself,
while $\alpha$ provides a stable, compact summary of the relative molecular
and spatial weighting rather than a quantitative measure of biological
importance.


\section{Implementation and Reproducibility Details}
\label{app:reproducibility}

\paragraph{Mouse Brain preprocessing.}
We use the coronal mouse-brain 10$\times$ Visium dataset distributed through
Squidpy~\citep{palla2022squidpy}. A fixed 2,200-spot subset is selected before
model fitting with random seed 42 and used identically for all compared
methods. We split these observations into 1,650 training and 550 test spots
before fitting any gene-level preprocessing. Counts are normalized to 10,000
per spot and log1p transformed. The top 3,000 highly variable genes are
selected on the training split only; gene-wise standardization and PCA-128 are
also fitted on training spots only, and the resulting transforms are applied
to test spots without refitting. Leiden clustering at resolution 0.5 and
random seed 42 yields 10 evaluation domains.

\paragraph{Human Liver preprocessing.}
GSE240429 contains four 10$\times$ Visium slices from donor C73.
Training uses A1+B1+D1 and testing uses C1.
The union of the top 1,000 highly variable genes across the training slices
produces 2,625 genes after deduplication. Gene-wise standardization and PCA-128
are fitted on A1+B1+D1 only, and C1 is transformed using the fitted training
statistics. Spatial coordinates are used only within their native slice;
cross-slice spatial-kernel affinities are masked to zero.

\paragraph{Image features.}
H\&E patches are extracted at each spatial spot from the high-resolution tissue
image using the \texttt{tissue\_hires\_scalef} parameter from Space Ranger.
Single-scale experiments use $96\times96$ native-image crops; multi-scale
experiments additionally use $224\times224$ crops from the same spot center.
Each crop is resized to the PLIP image encoder input resolution and processed
with the normalization associated with the public pretrained PLIP image
processor. The PLIP backbone remains frozen, and its output features are
$\ell_2$ normalized before the trainable adapters. No test-time gene or spatial
information enters the image encoder.

\paragraph{Kernel construction.}
Both $\Kgene$ and $\Kspat$ are RBF kernels whose bandwidths are fixed using the
median pairwise-distance heuristic on training data. During training, the
biological kernel
\begin{equation}
\Kstar=\alpha\Kgene+(1-\alpha)\Kspat
\end{equation}
is recomputed within each minibatch using the current learned $\alpha$. Self
entries are removed before selecting top-$k$ biological neighbors. In Human
Liver, $\Kspat(i,j)=0$ for pairs from different slices, so slice-local
coordinates are never interpreted as cross-slice physical proximity.

Evaluation is deliberately decoupled from the learned training weight. Every
model is evaluated with the same fixed kernel
\begin{equation}
\Keval=0.5\Kgene+0.5\Kspat,
\end{equation}
constructed independently of model predictions. The exact paired/self index is
removed before defining Bio-mAP, BioR, GeneR, and SpatR positive sets.

\paragraph{Optimization.}
All experiments are implemented in PyTorch and trained on a single NVIDIA
A100 40\,GB GPU.
We use AdamW with weight decay $10^{-4}$,
$\beta_1=0.9$,
$\beta_2=0.999$,
initial learning rate $3\times10^{-4}$,
cosine annealing,
batch size 256,
and 60 training epochs without early stopping.

We parameterize the constrained mixture weights as
$\alpha=\sigma(a)$, $\rho=\sigma(r)$, and $w_s=\sigma(u)$.
The unconstrained parameters $a$, $r$, and $u$, together with $\tau$,
$\eta_x$, and $\eta_g$, are initialized as
$\mathrm{logit}(0.6)$, $0$, $0$, $\log(0.07)$, $0.1$, and $0.1$,
respectively.

Loss weights are fixed to
\begin{equation}
\lambda_s=0.3,
\qquad
\lambda_g=0.1,
\qquad
\lambda_\ell=0.5.
\end{equation}
The reported $k$ and loss weights are selected on held-out validation data from
the training split and are fixed before final test evaluation. The loss
implementation uses mean normalization over anchors or valid pairwise entries,
matching the equations in Appendix~\ref{app:method}; in particular,
$\Lsoft$ is averaged over the $B$ anchors and $\Lglob$ over the $B^2$ kernel
entries.

All stochastic experiments report mean$\pm$standard deviation over five
random seeds.

\paragraph{Baseline implementation.}
CCA uses 50 components and at most 1,000 iterations. Ridge regression uses
regularization coefficient 1.0. The reported PLIP zero-shot result is a naive
unaligned lower bound: PCA-128 is fitted to training PLIP features and compared
by cosine similarity with the gene PCA representation without learned
cross-modal alignment. PLIP linear trains one linear projection for each
modality with exact bidirectional InfoNCE using the same optimizer and schedule
as BioKERN. BLEEP uses its original two-layer MLP encoder and
similarity-smoothed contrastive objective. BLEEP$^*$ replaces the BLEEP
projection architecture with the ResidualAdapter backbone while retaining the
BLEEP objective. Ret-only, Rank, and Shuffled use the same ResidualAdapter
architecture and differ only in the applied loss.


\section{Limitations and Broader Directions}
\label{app:limitations}

The current evaluation is restricted to two single-donor spatial
transcriptomics benchmarks. Human Liver evaluates cross-slice transfer within
one donor, while Mouse Brain uses a within-section split; neither setting
captures the full variation expected across donors, disease states,
sample-processing protocols, or spatial-transcriptomics platforms.

Biological neighborhoods are operationalized through transcriptomic similarity
and spatial proximity rather than external expert annotations. Bio-mAP is
therefore evaluated with a fixed, model-independent molecular--spatial kernel
and should still be interpreted together with transcriptomic, spatial,
cluster, and exact-pair metrics. The training-time mixture parameter $\alpha$
is a single scalar for each trained model; although it is stable to
initialization, it cannot represent region-specific changes in which
biological signal is most informative.

Larger-scale validation will be necessary to determine whether the learned
molecular--spatial weighting transfers across tissues and datasets. Future work
will therefore evaluate BioKERN on resources such as
HEST-1k~\citep{jaume2024hest}, study multi-donor and disease-stratified
settings, and incorporate richer priors such as cell-type relationships,
morphological neighborhoods, pathway activity, and cell--cell interaction
structure. The formulation is also compatible with modalities beyond gene
expression and histology, including spatial proteomics and multiplexed
imaging.

\end{document}